%% file: main.tex
\pdfoutput=1
\documentclass{article}
\usepackage[utf8]{inputenc}
\usepackage{graphicx}
\usepackage{caption}
\usepackage{subcaption}
\usepackage{amsmath}
\usepackage{amssymb}
\usepackage{hyperref}
\usepackage{natbib}

\title{Fair Classification under Covariate Shift and Missing Protected Attribute \\
        \large
            - an Investigation using Related Features}
\author{Manan Singh} 
\date{November 2021}

\begin{document}

\maketitle

% Abstract
\begin{center}
    \textbf{Abstract}
\end{center}
This study investigated the problem of fair classification under Covariate
Shift and missing protected attribute using a simple approach based on the
use of importance-weights to handle covariate-shift and,
\textit{Related Features} \cite{zhao2021you} to handle missing protected attribute.
% , and found that this approach was not significantly better than
% even simple baselines such as altogether removal of Related features.

\input{introduction}

\input{problem_setting}

\input{methodology}

\input{experiments}

\input{results}

\input{conclusion}

\bibliography{my_refs}
\bibliographystyle{plainnat}

\pagebreak
\appendix
\input{appendix_related_work}
\input{appendix_results.tex}

\end{document}

%% file: introduction.tex
\section{Introduction}

% Para-1: Ensuring fair classification is IMPORTANT.
Automated decision making has become prevalent in our society today.
Organizations such as companies, judiciaries, institutions, etc., rely on machine learning models, especially binary classifiers, to make important decisions about people, such as, loan crediting, admissions into university, etc.
These decisions can get biased towards a specific group, such as, race or gender, if the classifier has been trained using a biased training-set.
\footnote{A \textit{bias} in the binary-classification training-set can refer to, either an imbalance in the sample-sizes of the two groups (e.g., 1000 samples being available for males, whereas, only 100 being available for females), or an imbalance in the class-labels for the two groups (e.g., 90\% of males being labeled as positive, whereas, only 10\% of females being labeled as positive.)}
The field of fair-classification has developed many methods to tackle this problem, but mostly, they depend on the availability of the group-information, also known as the sensitive or the protected attribute.

% Para-2: But, the availability of true protected attribute, which most classification mechanisms require, might not be possible in a real-world setting. 
But the availability of the protected attribute might not be realistic in a real-world setting,
for example, due to a law prohibiting the collection of sensitive group information
such as gender or race, thus, making achieving fairness challenging.
To achieve fairness in such a case, when the true sensitive attribute is not
available, \cite{zhao2021you} have suggested using a set of non-sensitive features that have been identified by domain experts as being highly correlated with the missing true attribute and provided beforehand.
They call these as \textit{Related Features}.
We found that the use of related features to achieve fairness has been a quite unexplored direction in the fairness literature, and hence, explored their use to achieve fair classification under Covariate Shift.

% PARA-3: Covariate Shift is an important issue.
Covariate-Shift is a very realistic issue faced by machine learning models, when after being trained on a particular data distribution, they have to be used on a new and different data distribution.
We illustrate the issue by a hypothetical scenario of a company that had been using a model trained on the customer records of its own country for a past few years, but now,
wants to expand its market to another country.
As a result, it might face the difficulty of distribution-change and its existing model might prove unreliable.
Moreover, due to limitations of data collection, the company may find it difficult
to get labelled data for the new country that it could have used to train a new model.
Thus, to build an accurate classifier will be a challenge.
This is the standard problem of unsupervised covariate shift.

% PARA-4: Covariate Shift + Fairness
Further, the laws of the new country might require the company's classifier to also be fair on the country's population, but the knowledge of true protected attribute might not be available to the company due to limitations of data collection or legal restrictions.
In such a case, the company might decide to seek help of domain experts, identify a set of non-sensitive but related attributes, and use them to achieve fairness instead.

In such a novel hybrid setting that involves both missing protected attribute and covariate shift, we investigated a simple possible approach based on Related Features to address the joint issue.

%% file: problem_setting.tex
\section{Problem Setting}

We have been given $N_S$ labelled samples from the source domain  $ (\mathbf{X}^S, Y^S) =  \{(\mathbf{x}_i^S, y_i^S)\}_{i=1}^{N_S}$, and $N_T$ unlabelled samples $ \mathbf{X}^T = \{  \mathbf{x}_i^T \}_{i=1}^{N_T}$ from the target domain, where $\mathbf{x}_i \in \mathbb{R}^d$ and $y_i \in \{0, 1\}$.
Under covariate-shift assumption, the source-domain features $\mathbf{x}_i^S$ are assumed to be drawn from a source-distribution, $p_S(\mathbf{x})$, whereas, the target-domain features, $\mathbf{x}_i^T$, are assumed to be drawn from a different target-distribution $p_T(\mathbf{x})$. 
Conditional distribution $p(y|\mathbf{x})$ is assumed to be same in both the domains.

We wish to assign labels $ Y^T = \{y_i^T\}_{i=1}^{N_T}$ to the samples in the target-domain using a classifier, $g_{\theta}: \mathbf{X} \rightarrow Y$
\footnote{We will denote the classifier predictions after sigmoid with $h_{\theta}: \mathbf{X} \rightarrow [0, 1]$, or $\tilde{y}$. },
that is not only accurate, but also fair with respect to the missing protected attribute in the target-domain, $ A^T = \{(a_i^T)\}_{i=1}^{N_T}$.

Although the \textit{true} protected attribute is absent, a set of $K$ features, $F^T = \{ ( x_{i1}^T, x_{i2}^T, ..., x_{iK}^T )_{i=1}^{N_T} \}$ have been identified by domain-experts as being highly correlated with the true attribute, and have to be used to enforce fairness in the classifier.

%% file: methodology.tex
\section{Methodology}

\subsection{Handling Covariate Shift}
A standard way to build a classifier $h_{\theta}$ that is robust to covariate-shift involves weighting those source-domain samples that are similar to the target-domain samples by modifying the traditional classifier loss as follows:

\begin{equation}
    \mathcal{L}(\mathbf{X}^S, Y^S; \theta) =  \frac{1}{N} \sum_{i=1}^N
    w(\mathbf{x}_i^S) \;
    \mathcal{L}_{clf} (h_{\theta}(\mathbf{x}_i^S), y_i^S)
\end{equation}

where, $\mathcal{L}_{clf}(.)$ is the standard binary cross-entropy loss:

\begin{equation}
    \mathcal{L}_{clf} (h_{\theta}(\mathbf{x}_i^S), y_i^S) = -y_i^S \log h_{\theta}(\mathbf{x}_i^S)
    -(1 - y_i^S) \log (1 - h_{\theta}(\mathbf{x}_i^S) )
\end{equation}

and, $w(.)$ assigns the importance-weights to the source-domain samples.

\begin{equation}
    w(\mathbf{x}_i^S) = \frac{p_T (\mathbf{x}_i^S)}{p_S (\mathbf{x}_i^S)}
\end{equation}

The above density-ratio is commonly estimated using a classifier, such as logistic regression, trained to distinguish between source and target-domain features - $X_S$ and $X_T$ \cite{sugiyama2012density}.

\subsection{Fairness using Related Features}

A common approach to build a fair classifier is to add a fairness loss to the conventional classification loss.
When the true protected attribute is available,
\cite{zafar2017fairness} has suggested to use the correlation between this attribute, $A$, and predictions, $\tilde{Y} = h_{\theta}(\mathbf{X})$, as the fairness loss.

\begin{equation}
    \mathcal{L}_{fair}(A, \mathbf{X}; \theta) =
    \frac{1}{N}
    \left|
    \sum_{i=1}^N
    (a_i - \mu_a)
    (h_{\theta}(\mathbf{x}_i) - \mu_{h_{\theta}})
    \right|
\end{equation}

In the absence of true protected attribute but when $K$ related attributes $F = \{X_1, X_2, ..., X_K\}$ are available, \cite{zhao2021you} have extended the above fairness loss as follows:

\begin{equation}
    \mathcal{L}_{fair}(\{X_1, X_2, ..., X_K\}, \mathbf{X}; \theta) =
    \sum_{k=1}^K
    \lambda_k .
    \mathcal{L}_{fair}(X_k, \mathbf{X}; \theta)
\end{equation}
where, $\lambda_k$ is the feature-specific weight, and can be decided by domain-experts, set to uniform, or even learnt automatically using a scheme provided by \cite{zhao2021you}.

\subsection{A Hybrid Method}
As we aimed to jointly handle both covariate shift, and fairness in the absence of true protected attribute but in the presence of related attributes, we chose to minimize the following loss that incorporates both of the approaches mentioned in the previous subsections.

\begin{multline}
    \mathcal{L}(\mathbf{X}^S, Y^S, F^T) =\{X_1^T, ..., X_K^T\}, \mathbf{X}^T; \theta) =
    \frac{1}{N_S}
    \sum_{i=1}^{N_S}
    w_i^S .
    \mathcal{L}_{clf}(\mathbf{x}_i^S, y_i^S; \theta)
    \\ +
    \eta .
    \sum_{k=1}^K
    \lambda_k .
    \mathcal{L}_{fair}(X_k^T, \mathbf{X}^T; \theta)
\end{multline}
where, $\eta$ is a fairness coefficient and can be tuned as a hyper-parameter.

The former term was included to enforce the classifier to be accurate under covariate shift, while the latter term was included to achieve fairness in the target domain using related features.

%% file: experiments.tex
\section{Experiments}

\subsubsection*{Approaches}
To investigate the effectiveness of the hybrid approach, we ran experiments for
the following approaches:
\begin{enumerate}
      \item \textbf{Vanilla}: A simple MLP classifier trained with the usual binary
            cross-entropy loss.
      \item \textbf{Related Features removed}: Same as the vanilla approach, but
            the related features are removed from the feature-set.
      \item \textbf{Covariate-Shift Adapted}: An MLP classifier trained with a
            weighted binary-cross entropy loss, where the weights are covariate-shift importance
            weights.
      \item \textbf{Fair using Related Features}: An MLP classifier
            trained with the usual binary cross-entropy loss, plus another loss term
            to incorporate fairness using related features.
            \footnote{Uniform weights were used for the related features.}
      \item \textbf{Hybrid}: An MLP classifier trained with the covariate-shift weighted
            classifier loss, plus fairness loss using related features.
\end{enumerate}

In all the experiments, the source-domain samples were used for training, whereas,
the target-domain samples were used for validation and testing.

\subsubsection*{Experiment-Settings}
All the approaches used a three-layer Multi-layer Perceptron (MLP) having two
hidden layers of size 64 and 32 respectively.
The learning-rate was set to 0.01, and the batch-size to 256.
Early stopping based on total validation loss was used to stop the training.
\footnote{\textit{Early Stopping: } As the training proceeds over the epochs,
      the best seen (i.e. the minimum) total validation loss is kept track of.
      If the validation loss does not improve after `k' consequent epochs, the training
      is halted.
      `k' is called as the patience criteria.
      In our experiments, a patience of 5 was used.
}
For the approaches that involve a fairness coefficient, experiments were run for
the fairness coefficient range of [1e-5, 1e-4, 0.001, 0.01, 0.1, 1.0, 10, 100, 1000, 1e4, 1e5],
and for each value, the tradeoff between performance and fairness reported.

\subsubsection*{Datasets}

The experiments were run for the following datasets.
\begin{itemize}
      \item \textbf{ADULT}
            \footnote{\url{https://archive.ics.uci.edu/ml/datasets/adult}}:
            % CONTENT
            % -------
            % about - part of US Census 1994 database
            % binary classification. income >50K per year or less.
            % samples ~ 40 K
            % features ~ 12 (excluding race and sex). e.g., age, education, hours-per-week, etc.
            % sensitive attr.: Sex 
            % (note: sensitive attr. is not part of the feature-set)
            % related features: the 3 most correlated ones (i.e. relationship, marital-status, hours_per_week)
            % cov. shift: median of age
            This dataset is part of the US Census 1994 surveys, and contains around 45K records of people with 12 features such as age, education, hours-per-week, etc.
            \footnote{Sensitive features viz. race and gender are not counted.}
            The true protected attribute, $A$,  is considered to be sex, and the
            %three most correlated 
            features - relationship, hours-per-week, and marital-status are chosen to comprise the related feature-set $F$.
            For covariate-shift, the samples having age less than the median age are taken to form the source-domain, and the rest form the target-domain.

      \item \textbf{MEPS (Medical Expenditure Panel Survey}
            \footnote{\url{https://meps.ahrq.gov/mepsweb/data_stats/download_data_files_detail.jsp?cboPufNumber=HC-181}}
            \footnote{\url{https://github.com/Trusted-AI/AIX360/blob/master/aix360/data/meps_data/README.md}}:
            % CONTENT
            % -------
            % a large scale 2015 survey of US population that collected demographics, healthcare expenditure, and other self-reported medical information. 
            % binary classification. binarized the annual healthcare expenditure (above and below the median).
            % samples ~ 8K
            % features ~ 42 (excluding race and sex).
            % sensitive attr.: Sex 
            % (note: sensitive attr. is not part of the feature-set)
            % related features: the 3 most correlated ones (i.e. PREGNT_31, HONRD31, INCOME_M)
            % cov. shift: median of annual income
            This dataset contains around 8K records from a large scale 2015 survey of US population.
            Around 42 features had been collected that include information such as, demographics, healthcare expenditure, and other self-reported medical information.
            We have binarized the healthcare expenditure as low-cost or high-cost based upon it being less than the median value, or otherwise.
            Sex is considered as the true protected attribute, and the features -
            PREGNT\_31, HONRD31, and INCOME\_M form the set of related features $F$.
            \footnote{The full codebook containing the description of the feature-names can be found at \url{https://meps.ahrq.gov/data_stats/download_data_files_codebook.jsp?PUFId=H181}}
            For covariate-shift, the samples having annual income less than median are considered as the source-domain, and the rest form the target-domain.
\end{itemize}
Note the choice of the features to comprise the set of related features, $F$, was assumed to be part of \textit{domain knowledge}.

\subsubsection*{Evaluation Metrics}

To evaluate the classifier's performance, we reported the \textbf{AUC} score, and
to measure its fairness, we reported \textbf{Demographic Parity distance} ($\Delta_{DP}$):

$$\Delta_{DP}  = | \mathbb{E}( \hat{y} | a=1 ) -  \mathbb{E}( \hat{y} | a=0 ) |$$
where, $\hat{y}$ denotes the predicted label, and $a$ denotes the true protected attribute.
Note that this fairness metric relies only on the predictions, and not the actual labels $y$.

To report the results for an experiment, the experiment was repeated 5 times
with different seeds, and the test-set metrics averaged over these 5 runs reported.
%along with the standard deviations.

%% file: results.tex
\section{Results and Discussion}

\begin{figure}
      \centering
      \begin{subfigure}{0.8\textwidth}
            \centering
            \includegraphics[width=\textwidth]{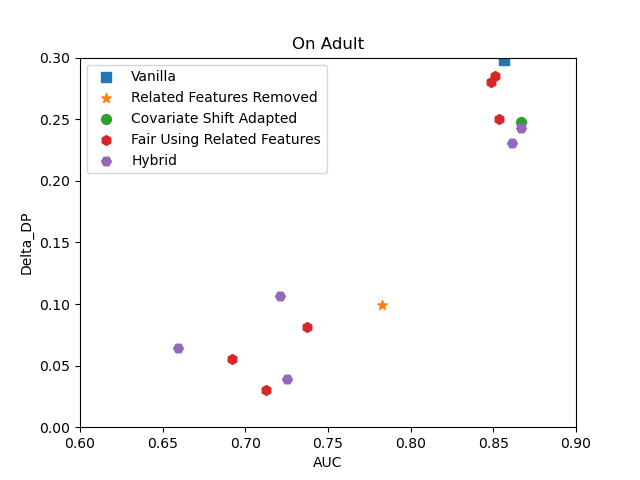}
      \end{subfigure}
      \begin{subfigure}{0.8\textwidth}
            \centering
            \includegraphics[width=\textwidth]{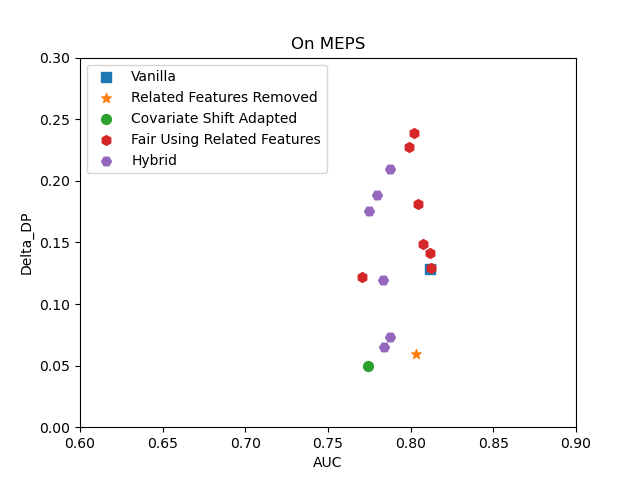}
      \end{subfigure}
      \caption{Performance v/s Fairness Tradeoffs for various approaches}
      \label{figure-tradeoff}
\end{figure}

For all the approaches, the performance-vs-fairness tradeoffs are plotted in Figure \ref{figure-tradeoff}.
(For the approaches involving fairness coefficient, the results obtained
for various values of fairness coefficient are plotted.)
For detailed results, see the appendix.

Our goal with trying the \textit{Hybrid} method, i.e. the one that uses related features
for fairness along with covariate shift adaptation, was to see if it can outperform
the baselines significantly either in terms of fairness, or performance, or both.
We made the following observations:
\begin{description}
      \item[Fairness] : In terms of best fairness,
            on the Adult dataset,
            it was a \textit{fair using related features} approach \cite{zhao2021you} that achieved
            the lowest $ \Delta_{DP} $ of around 0.05
            (Notice the lowest red dot in Figure \ref{figure-tradeoff}.),
            whereas,
            on the MEPS dataset,
            it was the \textit{covariate shift adapated} approach that achieved the lowest.
            (Notice the green dot near the bottom in Figure \ref{figure-tradeoff}.)
      \item[Performance] : In terms of best AUC, multiple approaches even other than the hybrid
            were able to achieve close to the best i.e. around 85\% on Adult, and around
            80\% on MEPS.
      \item[Fairness-Performance tradeoff] : In terms of the best tradeoff,
            both on the Adult and the MEPS datasets, the \textit{Related features removed} approach seemed
            to be the best, having, on Adult, an AUC of around 80\% and $ \Delta_{DP} $ of around 0.10; and
            on MEPS, an AUC of 80\% and $ \Delta_{DP} $ of 0.05.
            Although, for certain values of fairness coefficients, the \textit{Fair using related features} and \textit{Hybrid} approaches
            had similar tradeoffs, these did not seem significantly superior to the
            \textit{Related features removed} approach.
\end{description}

Thus, the overall results indicate that the hybrid method, i.e. the use of related
features along with covariate-shift adaptation, is not significantly
more efficient than even simple baselines such as altogether removal of the related features.

%% file: conclusion.tex
\section{Conclusion}

In this study, we investigated the problem of fair classification under covariate
shift and missing protected attribute using an approach involving importance
weights for covariate shift adaptation and fairness using related features
for handling the missing attribute, and found it to be not much significantly
efficient than even simple baseline such as removal of related features.

%% file: appendix_related_work.tex
\section{Related Work}

This work explored the problem of fair classification under covariate shift and
with missing protected attributes, and hence, lies at the intersection of the
areas - fair classification, covariate shift, and missing protected attribute.
We list the related works briefly below.

%theme: classic fair ml
Classic Fairness literature, that aims at achieving fair classification with respect
to a particular protected attribute, involves pre-processing (and feature representation learning)
\citep{zemel2013learning,louizos2015variational},
in-processing \citep{bechavod2017penalizing, agarwal2018reductions},
and post-processing \citep{hardt2016equality} techniques.

%theme: with Cov. shift.
To also handle the realistic problem of \textit{covariate-shift}, due to which
the data-distribution upon which the model has to be used (i.e. the target domain)
has changed from the distribution that was available during training
(i.e. the source domain), recent works aim at achieving fairness in the presence
of covariate-shift
\citep{singh2021fairness,rezaei2020robust,yoon2020joint,schumann2019transfer}.

But a huge limitation of such works is their reliance on a specific protected
attribute.
\cite{lahoti2020fairness, amini2019uncovering} have proposed solutions to achieve fairness when the protected attribute are hidden and never available, but their methods do not handle covariate shift.
\cite{coston2019fair} also handles covariate shift, but it assumes the availability
of the protected attribute either in the source or in the target domain.
We could not find any work in the literature that handles the complete
unavailability i.e. both from the source and target domains.

%% file: appendix_results.tex
\section{Detailed Results}

\begin{figure}[h]
      \centering
      \begin{subfigure}{\textwidth}
            \centering
            \includegraphics[width=0.8\textwidth]{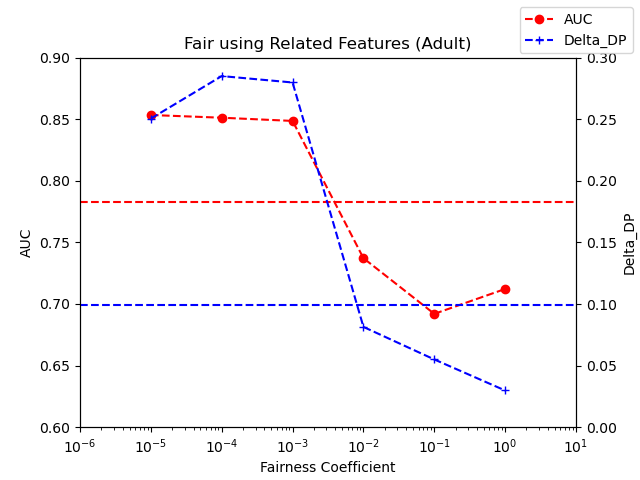}
            \caption{Fair using Related Features}
      \end{subfigure}
      \par\bigskip
      \begin{subfigure}{\textwidth}
            \centering
            \includegraphics[width=0.8\textwidth]{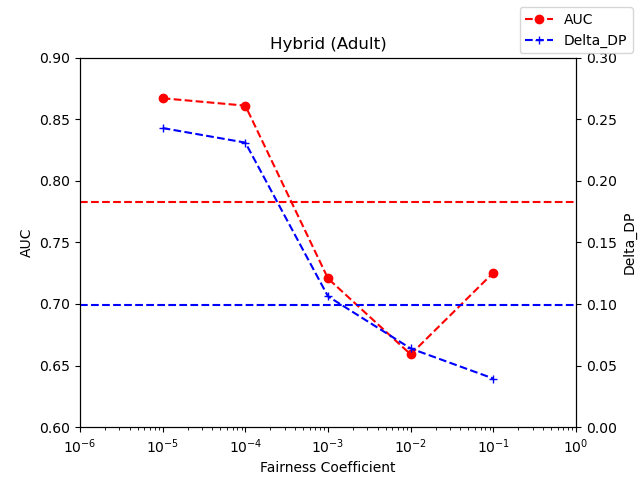}
            \caption{Hybrid (Covariate Shift Adapted + Fair using Related Features)}
      \end{subfigure}
      \caption{AUC v/s Fairness tradeoffs on Adult dataset (The thresholds correspond to the results of the "Related Features removed" approach.)}
      \label{figure-adult}
\end{figure}

\begin{figure}[h]
      \centering
      \begin{subfigure}{\textwidth}
            \centering
            \includegraphics[width=0.8\textwidth]{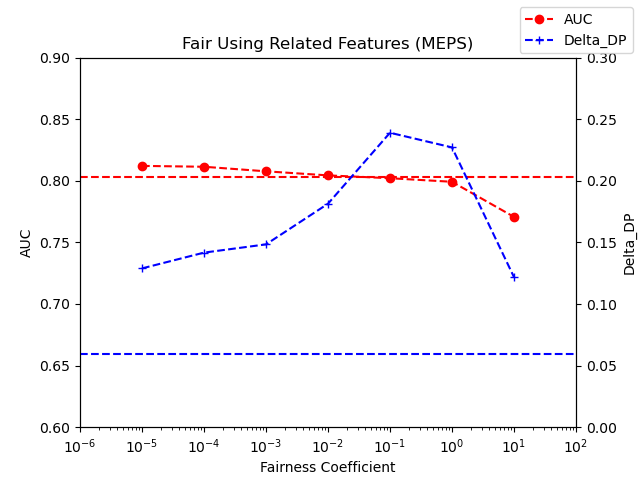}
            \caption{Fair using Related Features}
      \end{subfigure}
      \par\bigskip
      \begin{subfigure}{\textwidth}
            \centering
            \includegraphics[width=0.8\textwidth]{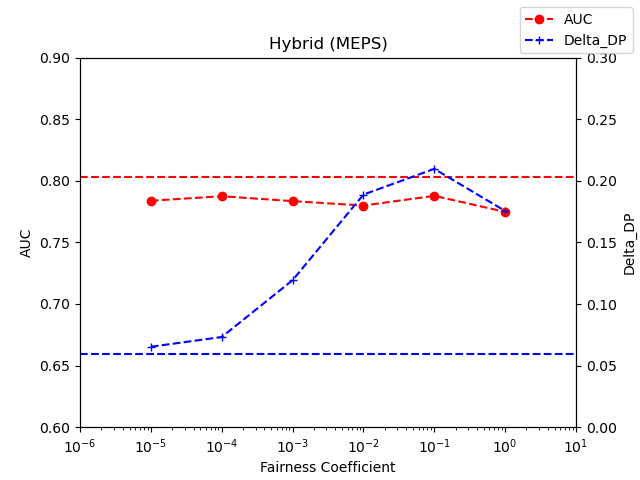}
            \caption{Hybrid (Covariate Shift Adapted + Fair using Related Features)}
      \end{subfigure}
      \caption{AUC v/s Fairness tradeoffs on MEPS dataset (The thresholds correspond to the results of the "Related Features removed" approach.)}
      \label{figure-meps}
\end{figure}

For the approaches that do not incorporate fairness, the results
are tabulated in Table \ref{table-adult} and \ref{table-meps}.
For the approaches that do incorporate fairness, the plots of the performance-versus-fairness tradeoffs over a varying
range of the fairness coefficient are displayed in Figure \ref{figure-adult} and \ref{figure-meps},
and the results tabulated in Tables
\ref{table-adult-fairness-rf},
\ref{table-adult-fairness-hybrid},
\ref{table-meps-fairness-rf} and
\ref{table-meps-fairness-hybrid}.
\footnote{Although, the fairness-coefficient-range of [1e-6, 1e6] was experimented with,
      only those fairness-coefficients that yielded non-zero $\Delta_{DP}$s are reported,
      because, the ones that led to zero $\Delta_{DP}$ led to very poor AUC values (of almost 50\%
      or below).}

The thresholds in the plots correspond to the results of the ``Related Features removed" approach,
and, in none of the plots could we find if the ``Hybrid'' approach had a significantly
better AUC v/s $\Delta_{DP}$ tradeoff.
\footnote{i.e. as compared with the tradeoff of ``Related Features removed", which was
      around 78\% (AUC) and 0.1 ($\Delta_{DP}$) on Adult; and
      around 80\% (AUC) and 0.05 ($\Delta_{DP}$) on MEPS.}

% Table of results (for non-fairness approaches)
\begin{table}[h]
      \centering
      \caption{Results on Adult dataset}
      \begin{tabular}{| l | c | c |}
            \hline
            \textbf{Method}          & \textbf{AUC} & \textbf{$\Delta_{DP}$} \\
            \hline
            Vanilla                  & 0.8565       & 0.2978                 \\
            Related Features Removed & 0.7827       & 0.0995                 \\
            Covariate Shift Adapted  & 0.8667       & 0.2476                 \\
            \hline
      \end{tabular}
      \label{table-adult}
\end{table}

\begin{table}[h]
      \centering
      \caption{Results on MEPS dataset}
      \begin{tabular}{| l | c | c |}
            \hline
            \textbf{Method}          & \textbf{AUC} & \textbf{$\Delta_{DP}$} \\
            \hline
            Vanilla                  & 0.8115       & 0.1281                 \\
            Related Features Removed & 0.8033       & 0.0596                 \\
            Covariate Shift Adapted  & 0.7742       & 0.0496                 \\
            \hline
      \end{tabular}
      \label{table-meps}
\end{table}

% Table of results (for fairness approach - fair using related features, and adult)
\begin{table}[h]
      \centering
      \caption{Results for a varying range of fairness coefficient -
            Approach: Fair using Related Features; Dataset: Adult.}
      \begin{tabular}{| c | p{2cm} | c | c |}
            \hline
            \textbf{Sn.} & \textbf{Fairness Coefficient} & \textbf{AUC} & \textbf{$\Delta_{DP}$} \\
            \hline
            1            & 1.00E-05                      & 0.8534       & 0.2505                 \\
            2            & 0.0001                        & 0.8512       & 0.285                  \\
            3            & 0.001                         & 0.8486       & 0.2798                 \\
            4            & 0.01                          & 0.7372       & 0.0814                 \\
            5            & 0.1                           & 0.692        & 0.055                  \\
            6            & 1                             & 0.7122       & 0.0298                 \\
            \hline
      \end{tabular}
      \label{table-adult-fairness-rf}
\end{table}

% Table of results (for fairness approach - hybrid, and adult)
\begin{table}[h]
      \centering
      \caption{Results for a varying range of fairness coefficient -
            Approach: Hybrid; Dataset: Adult.}
      \begin{tabular}{| c | p{2cm} | c | c |}
            \hline
            \textbf{Sn.} & \textbf{Fairness Coefficient} & \textbf{AUC} & \textbf{$\Delta_{DP}$} \\
            \hline
            1            & 1.00E-05                      & 0.867        & 0.2428                 \\
            2            & 0.0001                        & 0.861        & 0.231                  \\
            3            & 0.001                         & 0.7207       & 0.1062                 \\
            4            & 0.01                          & 0.6591       & 0.0639                 \\
            5            & 0.1                           & 0.7254       & 0.0395                 \\
            \hline
      \end{tabular}
      \label{table-adult-fairness-hybrid}
\end{table}

% Table of results (for fairness approach - fair using related features, and meps)
\begin{table}[h]
      \centering
      \caption{Results for a varying range of fairness coefficient -
            Approach: Fair using Related Features; Dataset: MEPS.}
      \begin{tabular}{| c | p{2cm} | c | c |}
            \hline
            \textbf{Sn.} & \textbf{Fairness Coefficient} & \textbf{AUC} & \textbf{$\Delta_{DP}$} \\
            \hline
            1            & 1.00E-05                      & 0.8121       & 0.1289                 \\
            2            & 0.0001                        & 0.8114       & 0.1417                 \\
            3            & 0.001                         & 0.8077       & 0.1484                 \\
            4            & 0.01                          & 0.8044       & 0.1814                 \\
            5            & 0.1                           & 0.8021       & 0.239                  \\
            6            & 1                             & 0.7992       & 0.2271                 \\
            7            & 10                            & 0.7707       & 0.1218                 \\
            \hline
      \end{tabular}
      \label{table-meps-fairness-rf}
\end{table}

% Table of results (for fairness approach - hybrid, and meps)
\begin{table}[h]
      \centering
      \caption{Results for a varying range of fairness coefficient -
            Approach: Hybrid; Dataset: MEPS.}
      \begin{tabular}{| c | p{2cm} | c | c |}
            \hline
            \textbf{Sn.} & \textbf{Fairness Coefficient} & \textbf{AUC} & \textbf{$\Delta_{DP}$} \\
            \hline
            1            & 1.00E-05                      & 0.7838       & 0.0653                 \\
            2            & 0.0001                        & 0.7875       & 0.0732                 \\
            3            & 0.001                         & 0.7835       & 0.1193                 \\
            4            & 0.01                          & 0.7799       & 0.1888                 \\
            5            & 0.1                           & 0.7878       & 0.2097                 \\
            6            & 1                             & 0.7747       & 0.1755                 \\
            \hline
      \end{tabular}
      \label{table-meps-fairness-hybrid}
\end{table}

% AUC-vs-Fairness tradeoff-plots

